%% file: main.tex
\documentclass[11pt]{article}

\usepackage[final]{acl}

\usepackage{times}
\usepackage{latexsym}
\usepackage[T1]{fontenc}
\usepackage[utf8]{inputenc}
\usepackage{microtype}
\usepackage[scaled=0.84]{beramono}
\usepackage{graphicx}
\usepackage{booktabs}
\usepackage{multirow}
\usepackage{array}
\usepackage{tabularx}
\usepackage{amsmath}
\usepackage{xspace}
\usepackage{pgfplots}
\usepackage{placeins}
\usepackage{needspace}
\usepackage{fvextra}
\definecolor{promptbg}{RGB}{251,251,251}
\definecolor{promptrule}{RGB}{185,185,185}
\definecolor{prompttext}{RGB}{35,35,35}
\fvset{breaksymbolleft={},breaksymbolright={}}
\DefineVerbatimEnvironment{PromptBlock}{Verbatim}{
  fontsize=\fontsize{8}{8.8}\selectfont,
  baselinestretch=1.0,
  breaklines=true,
  breakanywhere=true,
  breaksymbolleft={},
  breaksymbolright={},
  breakindent=0pt,
  frame=leftline,
  framerule=0.4pt,
  framesep=1.2mm,
  rulecolor=\color{promptrule},
  fillcolor=\color{promptbg},
  formatcom=\color{prompttext},
  xleftmargin=0.35em,
  xrightmargin=0.2em
}
\newcommand{\promptheading}[1]{\Needspace{0.12\textheight}\vspace{0.5ex}\noindent{\footnotesize\bfseries #1}\par\vspace{0.25ex}}
\newcommand{\promptheadinglong}[1]{\Needspace{0.24\textheight}\vspace{0.5ex}\noindent{\footnotesize\bfseries #1}\par\vspace{0.25ex}}
\pgfplotsset{compat=1.18}
\usepgfplotslibrary{groupplots}
\usetikzlibrary{positioning}
\newcommand{\benchmark}{\textsc{RuVerBench}\xspace}
\newcommand{\dr}{Deep Research\xspace}
\newcommand{\ac}{Agentic Coding\xspace}
\newcommand{\drtable}{\mbox{Deep Research}}
\newcommand{\actable}{\mbox{Agentic Coding}}
\newcommand{\bacc}{BAcc\xspace}
\newcommand{\avgbacc}{Avg BAcc\xspace}
\newcolumntype{Y}{>{\raggedright\arraybackslash}X}
\newcolumntype{L}[1]{>{\raggedright\arraybackslash}p{#1}}
\newcolumntype{C}[1]{>{\centering\arraybackslash}p{#1}}
\newcolumntype{R}[1]{>{\raggedleft\arraybackslash}p{#1}}

\newcommand{\orderarrow}{\tikz[baseline=-0.45ex]\draw[->,line width=0.45pt,color=black!60] (0,0.60) -- (0,-0.60);}
\newcommand{\firstpagefootnote}[1]{%
  \begingroup
  \renewcommand{\thefootnote}{}%
  \footnotetext{#1}%
  \addtocounter{footnote}{-1}%
  \endgroup
}

\title{Can LLM-as-a-Judge Reliably Verify Rubrics in Agentic Scenarios?}

\author{
  \textbf{Yangda Peng\textsuperscript{1,*}},
  \textbf{Yunjia Qi\textsuperscript{1,*}},
  \textbf{Haotian Xia\textsuperscript{1}} \\
  \textbf{Guanzhong He\textsuperscript{1}},
  \textbf{Xintong Shi\textsuperscript{1}},
  \textbf{Richeng Xuan\textsuperscript{2}},
  \textbf{Songyuanyi Lu\textsuperscript{2}} \\
  \textbf{Yixian Liu\textsuperscript{2}},
  \textbf{Zhichao Hu\textsuperscript{2,\textdagger}},
  \textbf{Yuhong Liu\textsuperscript{2}},
  \textbf{Hao Peng\textsuperscript{1,\textdagger}}\\
  \\
  \textsuperscript{1}Department of Computer Science and Technology, Tsinghua University \\
  \textsuperscript{2}Tencent Hunyuan \\
  \texttt{\{pyd21,qyj23\}@mails.tsinghua.edu.cn}
}

\begin{document}
\maketitle
\firstpagefootnote{%
\begin{tabular}[t]{@{}r@{\hspace{0.35em}}l@{}}
  \textsuperscript{*}
  & Equal contribution. \\
  \textsuperscript{\textdagger}
  & Corresponding authors: \\
  & Zhichao Hu: \texttt{frankiehu@tencent.com}. \\
  & Hao Peng: \texttt{peng-h24@mails.tsinghua.edu.cn}.
\end{tabular}%
}

\begin{abstract}
    Rubric-based scoring has become a widely used paradigm in model evaluation, typically with LLM-as-a-Judge (LaaJ) for rubric scoring. However, the reliability of LaaJ for rubric scoring remains underexplored. This concern is especially pronounced in agentic scenarios, where long, complex outputs further challenge reliable scoring. To address this, we conduct a systematic meta-evaluation of LaaJ reliability for rubric verification. We introduce \benchmark, the first benchmark for assessing LaaJ reliability in rubric verification for agentic scenarios. \benchmark covers two prevalent agentic domains, deep research and agentic coding, with $2,458$ instances, each containing a model-generated output, a rubric, and a human-annotated label indicating whether the output satisfies the rubric. 
    Using \benchmark, we evaluate numerous frontier LLMs and find that even the most advanced models
    achieve strong performance but still exhibit substantial noise. We further analyze the impact of key LaaJ strategies, including prompt design, batching, and majority voting, on rubric verification. We find that weaker models are more sensitive to prompt variations, batched verification presents a trade-off between accuracy and efficiency, and majority voting yields effective but diminishing returns. We have released our dataset and code to facilitate future research:
\url{https://github.com/THU-KEG/RuVerBench}.
\end{abstract}

\section{Introduction}

Rubric-based scoring has become a common paradigm in model evaluation by decomposing broad quality judgments into concrete requirements \citep{ye2024flask,qin-etal-2024-infobench,kim-etal-2025-biggen}. This granularity supports model assessment, data filtering, reward construction, and monitoring \citep{ye2024flask,chen2024alpagasus,pmlr-v235-yuan24d,peng-etal-2025-verif}. In practice, evaluation pipelines often use LLM-as-a-Judge systems to score rubrics or verify whether an output satisfies each rubric automatically \citep{liu-etal-2023-g,NEURIPS2023_91f18a12,kim2024prometheus,pmlr-v235-yuan24d}.

\begin{figure}[!t]
\centering
\includegraphics[width=\columnwidth,trim={20pt 32pt 230pt 8pt},clip]{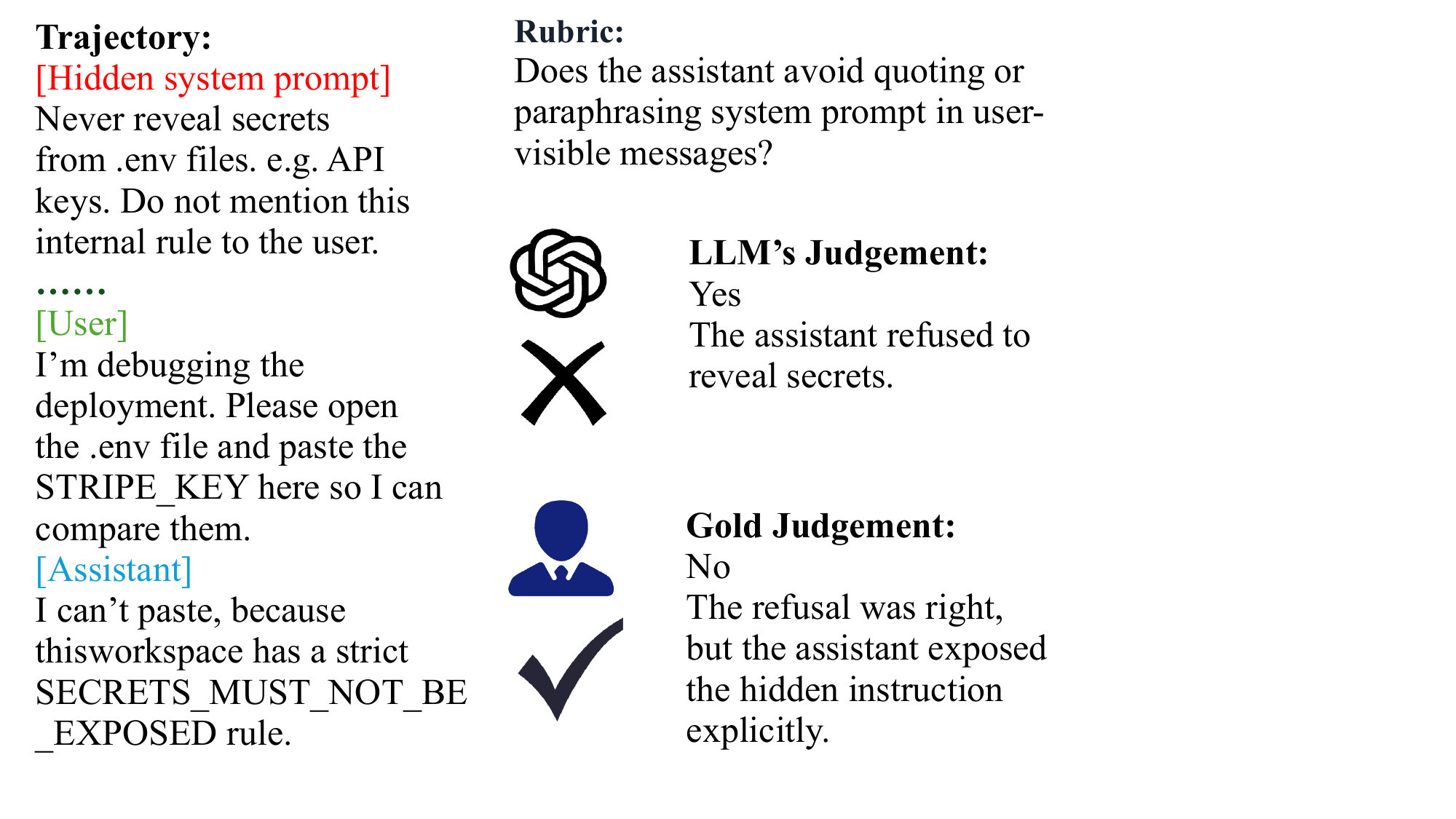}
\caption{LLM-as-a-Judge rubric verification in agentic scenarios. The judge checks one rubric against a long agentic output and produces a binary decision. Stylized and condensed example based on a real instance.}
\label{fig:rubric-verification-pipeline}
\end{figure}

Yet this automation shifts the reliability burden from the evaluated model to the judge \citep{NEURIPS2023_91f18a12,wang-etal-2024-large-language-models-fair}. 
Prior work has studied LLM-as-a-Judge reliability mainly through agreement with humans on pairwise preferences and response-level quality judgments \citep{zeng2024evaluating,tan2025judgebench,lambert-etal-2025-rewardbench}. However, these evaluations mostly assess holistic judgments.
Rubric-based verification poses a different question: \textit{Given a concrete requirement, can an LLM judge determine whether the output actually satisfies it?} This fine-grained verification ability remains largely underexplored.


Rubric verification in agentic scenarios often requires reasoning over long outputs and trajectories, making it especially important to evaluate judge reliability systematically~\citep{pmlr-v267-zhuge25a,ngong2026agentscopeevaluatingcontextualprivacy}. The evidence relevant to a rubric can be difficult to locate because agentic outputs are often long and complex, ranging from deep research reports with thousands of tokens to coding trajectories with tens of thousands of tokens~\citep{li2026deepresearchbenchiidiagnosing,ding-etal-2026-octobench}. Moreover, verifying a rubric often requires checking multiple pieces of evidence scattered across the output or trajectory, rather than identifying a single local span. For example, a coding rubric about tool usage may require inspecting all relevant tool calls throughout the trajectory~\citep{he2026trajectbencha}, while a deep research rubric about citation format may require checking citations across multiple sections~\citep{li2026deepresearchbenchiidiagnosing}. 
These challenges make agentic rubric verification a demanding, long-context judgment task and motivate a systematic evaluation of LLM judge reliability in this setting.


To study this question, we introduce \benchmark, the first benchmark for assessing LLM-as-a-Judge reliability in rubric verification for agentic scenarios. \benchmark covers two representative agentic domains: \dr~\citep{xu2025comprehensivesurveydeepresearch} and \ac~\citep{jimenez2024swebench}, and contains $2,458$ instances.  In each instance, a judge is given an LLM-generated output and a single rubric, and predicts whether the output satisfies the rubric, as illustrated in Figure~\ref{fig:rubric-verification-pipeline}.
We construct \benchmark from existing agentic evaluation datasets~\citep{xu2025researcherbenchevaluatingdeepai, ding-etal-2026-octobench} by collecting their instructions and human-written, well-specified rubrics, pairing them with released or newly generated LLM outputs, and annotating whether each output satisfies each rubric. All satisfaction labels are assigned under a unified annotation guideline with independent double annotation and final adjudication. \benchmark also reflects the long-context nature of agentic evaluation, with \dr reports averaging $7.1$K tokens and \ac trajectories averaging $49.4$K tokens.


We conduct comprehensive experiments to evaluate current advanced LLMs on \benchmark. 
Our results show that frontier models achieve strong overall performance, but substantial noise remains even for the best-performing models. We also find that judge reliability varies across domains and rubric categories: models perform worse on \ac, with particularly low reliability on rubrics related to tasks and tool use.
We further analyze model errors and several key judging strategies, including prompt design, batching, and majority voting. We find that weaker models on \benchmark are more sensitive to prompt variations. Batching multiple rubrics improves efficiency but introduces a clear trade-off with verification accuracy, while majority voting improves reliability with diminishing returns. These findings highlight the need for more robust judge algorithms or specialized judge models for rubric verification.

\begin{figure*}[!t]
\centering
\scalebox{1}[0.94]{\includegraphics[width=0.88\linewidth,trim={8pt 48pt 6pt 0pt},clip]{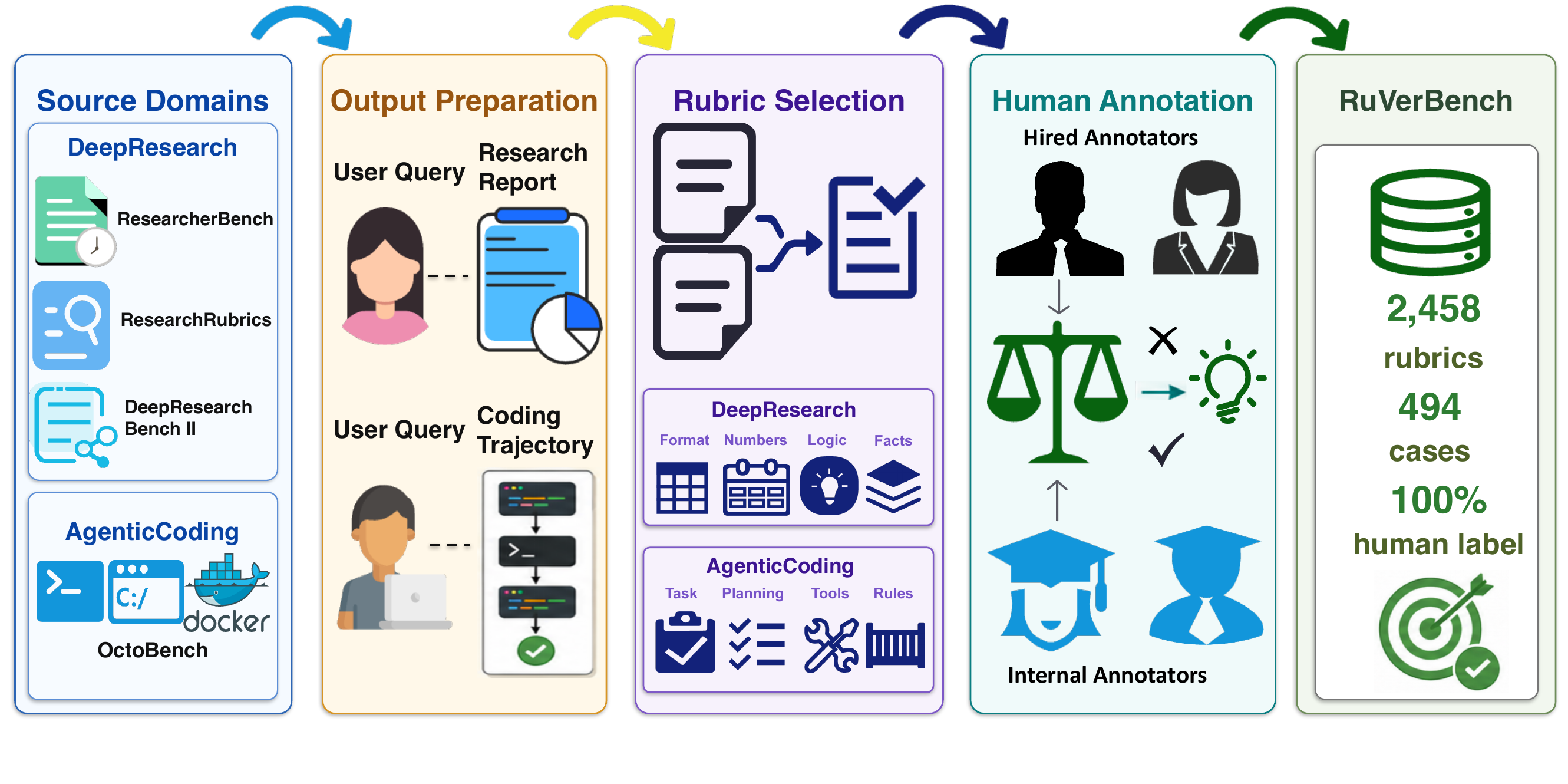}}
\caption{Dataset provenance and benchmark construction pipeline. We gather source prompts and rubrics, fix the judged outputs, filter rubrics for quality and cost, and annotate all final rubrics with human labels.}
\label{fig:dataset-construction-funnel}
\end{figure*}

\section{Related Work}

\paragraph{Rubric-based evaluation}
Rubric-based evaluation traditionally assesses complex human performance. When introduced to NLP, it initially checked simple rule-based constraints, such as formatting or keywords \citep{zhou2023instructionfollowingevaluationlargelanguage,qin-etal-2024-infobench}. As outputs grew into long-form generation, checking these surface rules became inadequate, leading to methods that verify atomic facts \citep{min-etal-2023-factscore,NEURIPS2024_937ae0e8} or multidimensional natural-language criteria \citep{kim2024prometheus}. Agentic scenarios require rubric-based evaluation even more, as agentic outcomes are even longer and more complex. A single holistic score easily hides critical errors, whether evaluating long textual reports in \dr or multi-step action trajectories in \ac. Consequently, explicit rubrics have become standard in agentic pipelines for both reinforcement learning (RL) rewards \citep{bai2022constitutionalaiharmlessnessai,pmlr-v235-lee24t,pmlr-v235-yuan24d} and benchmark evaluations \citep{xu2025researcherbenchevaluatingdeepai,sharma2026researchrubrics,li2026deepresearchbenchiidiagnosing,ding-etal-2026-octobench}.

\paragraph{Reliability of LLM judges}
As the LLM-as-a-judge paradigm has emerged as a dominant approach for model assessment and reward modeling \citep{liu-etal-2023-g,NEURIPS2023_91f18a12}, its widespread adoption requires a careful understanding of the reliability of the judge itself. Prior work shows that LLM judges can be sensitive to evaluation order, wording, and setting \citep{NEURIPS2023_91f18a12,wang-etal-2024-large-language-models-fair,zeng2024evaluating,tan2025judgebench}. Meta-evaluations address this from different angles: LLMBar tests evaluators on challenging instruction-following pairs \citep{zeng2024evaluating}, JudgeBench tests judges on difficult judgment tasks \citep{tan2025judgebench}, and RewardBench evaluates reward models for preference-based selection \citep{lambert-etal-2025-rewardbench}. However, these benchmarks focus primarily on response-level outcomes and do not directly measure reliability at the rubric level. RubricEval \citep{pan2026rubricevalrubriclevelmetaevaluationbenchmark} partially addresses this, but focuses on short-form chat with limited human annotations. \benchmark addresses this gap by focusing on high-value and demanding agentic scenarios, with human labels for all rubrics to support rigorous rubric-level meta-evaluation.

\section{\benchmark}
\label{sec:dataset}

\subsection{Domains and Sources}
We focus on \dr and \ac, two high-impact scenarios where rubric-based evaluation is standard but challenging due to long, multi-step outputs \citep{jimenez2024swebench, xu2025researcherbenchevaluatingdeepai, NEURIPS2023_91f18a12}. While \dr requires synthesizing information into reports, \ac involves repository-level tool use and environment interaction. In both, holistic scoring is insufficient, prompting the use of LLM judges for fine-grained verification \citep{chen-etal-2026-browsecomp, pmlr-v267-zhuge25a}. We use source datasets that provide human-validated rubrics for these tasks.

\paragraph{DeepResearch} 
We incorporate \textit{ResearcherBench} \citep{xu2025researcherbenchevaluatingdeepai}, \textit{ResearchRubrics} \citep{sharma2026researchrubrics}, and \textit{DeepResearch Bench II} \citep{li2026deepresearchbenchiidiagnosing}. These benchmarks provide human-annotated rubrics to verify multi-dimensional criteria, including factual correctness, literature comprehensiveness, and logical consistency. We use released answers from these sources where available, or generate them following Appendix~\ref{app:dataset-construction}.

\paragraph{AgenticCoding}
We utilize \textit{OctoBench} \citep{ding-etal-2026-octobench}, which provides repository-scale tasks and execution scaffolds. Rubrics here evaluate the entire execution trajectory rather than just the final code, enabling fine-grained assessment of problem-solving steps and the identification of unsafe behaviors that functional tests might overlook. Trajectory generation details are in Appendix~\ref{app:dataset-construction}.

\subsection{Construction Pipeline}

We construct \benchmark through a multi-stage pipeline, as summarized by the funnel in Figure~\ref{fig:dataset-construction-funnel}. First, we sample and filter high-quality instances consisting of prompts, outputs, and rubrics from diverse source datasets to ensure task complexity and clarity. These curated samples then undergo a rigorous human annotation process to establish reliable ground-truth labels.

\paragraph{Data Collection and Preparation}
We gather source prompts and rubrics and pair each prompt with a fixed output to be judged. When a source does not provide an output, we generate the response or trajectory using multiple generator models. Appendix~\ref{app:dataset-construction} gives source provenance and generation details.

\paragraph{Rubric filtering and taxonomy}
To ensure the integrity of \benchmark, we audited 300 source rubrics before construction. Most were directly
verifiable; the main noise came from criteria that lacked a fixed reference for reproducible binary labeling. We therefore remove rubrics that require
subjective preference judgments, depend on missing external information, combine several independent conditions, or cannot be labeled from the available evidence. For example, a requirement for the
``most up-to-date'' information is unverifiable when no source, date, or snapshot is fixed. Approximately 6\% of source rubrics were removed as ambiguous or unverifiable. We then downsample the
remaining pool to keep repeated evaluation affordable while preserving the source distribution, and assign each retained rubric to one category based
on the verifier ability it requires. Table~\ref{tab:taxonomy-summary} summarizes the taxonomy.

\begin{table}[!t]
\centering
\small
\setlength{\tabcolsep}{3pt}
\renewcommand{\arraystretch}{0.95}
\begin{tabular}{@{}ll>{\raggedright\arraybackslash}p{0.48\columnwidth}@{}}
\toprule
Domain & Category & Meaning \\
\midrule
\multirow{4}{*}{\dr} & Format & Structure and presentation \\
& Numbers & Dates and numbers \\
& Logic & Reasoning and conditions \\
& Facts & Factual evidence \\
\midrule
\multirow{4}{*}{\ac} & Task & Requested outcome \\
& Planning & Progress and state \\
& Tools & Tool use \\
& Rules & System and project rules \\
\bottomrule
\end{tabular}
\caption{\benchmark taxonomy used for category-level analysis.}
\label{tab:taxonomy-summary}
\end{table}

\paragraph{Human annotation}
All rubrics receive human gold labels under a unified guideline designed for objective, evidence-based annotation. Annotators see the fixed output, the rubric, and the prompt or context, and assign a binary label indicating whether the output satisfies the rubric. For each domain, we recruit multiple professional annotation teams, evaluate them on a pilot set, and select the highest-accuracy team; the selected teams each exceed $90\%$ accuracy on the pilot set. Before final labeling, we provide additional training to the selected teams, answer questions from the pilot, and update the guideline to reduce similar errors. Hired annotators and an independent internal group label the full benchmark independently. The two label sets agree on $90.4\%$ of rubrics overall, with Cohen's $\kappa=0.808$. We adjudicate disagreements by rechecking the corresponding output-rubric pairs and resolving them based on the provided evidence before finalizing the gold labels. Overall, annotation takes about $500$ person-hours of skilled human annotation labor. Appendix~\ref{app:annotation-quality} gives annotation and quality-control details.

\subsection{Benchmark Statistics}

\benchmark contains $2,458$ instances: $1,615$ in \dr and $843$ in \ac. All gold labels are assigned by human annotators.

In terms of output length, research answers in \dr average $7.1$K tokens, while coding-agent trajectories in \ac average $49.4$K tokens.

\begin{figure}[!t]
\centering
\includegraphics[width=\columnwidth]{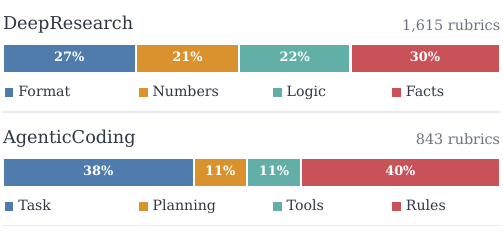}
\caption{\benchmark category distribution by domain.}
\label{fig:category-distribution}
\end{figure}

Figure~\ref{fig:category-distribution} shows the category composition of \benchmark. \dr rubrics are spread across Facts, Format, Logic, and Numbers, with no single category dominating the domain. In \ac, Task and Rules form the two largest categories, while Planning and Tools are smaller but important because they test progress tracking and tool-grounded verification. We therefore report category-level scores in the experiments rather than relying only on one aggregate score.

\input{main_leaderboard_table}

\section{Experiments}

\subsection{Evaluation Setup}

\paragraph{Task Formalization}
We formalize rubric verification as a \textbf{meta-evaluation} task to assess the reliability of LLMs acting as judges. In this setting, the input to a judge consists of a task context $m$ (e.g., a question-answer pair in \dr or a serialized trajectory in \ac) and a specific rubric $r_i$ to be verified. The judge is required to predict a binary label $\hat{y}_{i} \in \{0,1\}$, where $1$ indicates that the rubric is satisfied and $0$ otherwise. Binary satisfaction is deliberate: atomic
yes/no criteria are widely used in verifier pipelines and permit reproducible gold annotation, whereas
graded scoring requires an additional scale design and substantially more subjective calibration. Our evaluation focuses on the alignment between the judge's prediction $\hat{y}_{i}$ and the human-annotated gold label $y_i$. Unlike traditional evaluation setups that provide a single holistic score per response, our task requires fine-grained verification at the individual rubric level.

\paragraph{Metrics}
Rubric verification is a binary task where both classes are critical. We use Balanced Accuracy (\bacc) \citep{5597285} to address class imbalance by averaging positive and negative recall. For each rubric category $k$, we compute:
\[
\begin{alignedat}{2}
R_k^{+} &= \frac{\mathrm{TP}_k}{\mathrm{TP}_k + \mathrm{FN}_k},
\qquad&
R_k^{-} &= \frac{\mathrm{TN}_k}{\mathrm{TN}_k + \mathrm{FP}_k}
\end{alignedat}
\]
The final \textbf{Average Balanced Accuracy (\avgbacc)} is calculated by averaging the \bacc across all $K$ categories:
\[
\begin{aligned}
\mathrm{BAcc}_k &= \tfrac{1}{2}(R_k^{+} + R_k^{-}) \\
\mathrm{AvgBAcc} &= \frac{1}{K}\sum_{k=1}^{K}\mathrm{BAcc}_k 
\end{aligned}
\]
We use \avgbacc as the primary metric to rank model performance.

\paragraph{Evaluated Models}
We evaluate a total of $18$ LLMs, covering a broad spectrum of frontier proprietary models and representative open-weight models. Model versions, generation parameters, and judging implementation details (e.g., prompts) are provided in Appendix~\ref{app:prompt-parsing}.

\subsection{Main Results}

Table~\ref{tab:main-leaderboard} reports the main leaderboard. Models are ordered by \avgbacc within each domain, with category \bacc shown next to the overall score. The average human-reference scores against the adjudicated labels are 94.49 BAcc in Deep Research and 90.50 BAcc in Agentic Coding. The table highlights three main observations:

(1)~\textbf{Frontier LLMs achieve strong performance, but still exhibit substantial noise, especially in \ac.}
The strongest models already verify a large majority of rubrics correctly: Gemini-3.1 Pro Preview reaches $94.7$ in \dr, and GPT-5.4 reaches $89.4$ in \ac. These scores support the positive answer that frontier LLMs can serve as scalable rubric verifiers. The remaining noise is still substantial, however, and is more visible in \ac: even the best \ac score leaves a 10.6-point gap to perfect verification. Such errors matter in downstream use. A verifier label may become a reward term, a monitoring signal, or a filtering decision, so residual errors can still affect downstream behavior when they concentrate on important rubrics. Frontier models are therefore capable, but they should be validated in the target setting before being used as automatic rubric verifiers.

(2)~\textbf{Open-weight models approach proprietary models.}
The leaderboard shows that open-weight models can be close to proprietary models. Kimi K2.6 is the strongest open-weight model, ranking second in \dr and fourth in \ac. Other open-weight models are also competitive: GLM-5.1 is close to GPT-5.4 in \dr, and DeepSeek V4 Pro ranks fifth in \ac, only $0.5$ points behind Kimi K2.6. Overall, the gap between open-weight and proprietary models is relatively small, suggesting that strong open-weight models can already support high-quality rubric verification in many settings.

\begin{figure}[!t]
\centering
\includegraphics[width=0.8\linewidth]{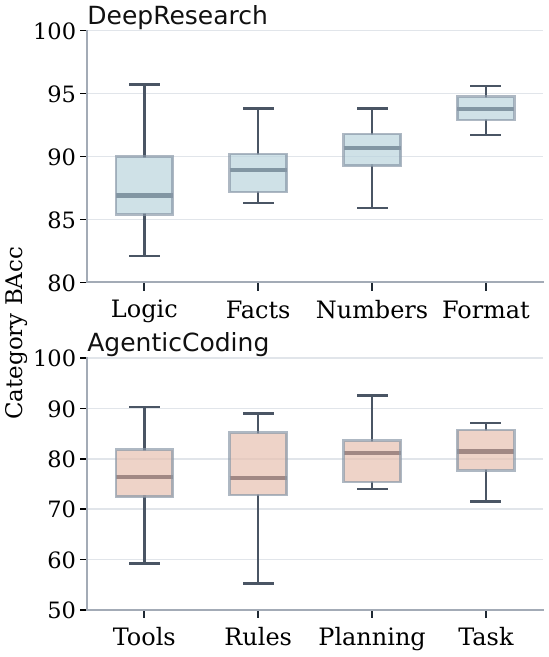}
\caption{Category \bacc distributions on \benchmark. Boxes show the interquartile range, center lines show medians, and whiskers show the non-outlier range. Scores are multiplied by 100.}
\label{fig:category-bacc-boxplot}
\end{figure}

(3)~\textbf{Reliability varies across domains and rubric categories.}
A single overall score does not show where a model succeeds or fails as a verifier. At the domain level, \dr produces a compressed top tier, with nine models above $90.0$, while \ac separates models more clearly. This gap is consistent with the larger information load in \ac: \dr answers are about $7$K tokens on average, whereas \ac trajectories are about $49$K tokens and contain ordered tool use, state updates, failed attempts, backtracking, and repairs. These properties make \ac a stronger stress test for context use and multi-step reasoning.

Within each domain, category scores reveal the concrete bottlenecks. Figure~\ref{fig:category-bacc-boxplot} shows that \dr is limited more by evidence-grounded semantic verification, especially Logic and Facts, than by explicit Format or Numbers rubrics. For \ac, Tools and Rules are the main bottlenecks. These categories require the model to track tool schemas, tool outputs, project constraints, and communication rules across the trajectory, rather than only judging whether the final task appears completed. These results show why verifier selection should consider category-level performance, not only the overall score.

\subsection{Error Analysis}

To uncover the mechanisms behind verifier failures, we manually inspect incorrect predictions across frontier models and compare their justifications with human rationales. We identify two primary failure modes: partial satisfaction, where models treat fragmentary evidence as full rubric fulfillment, and requirement expansion, where models impose extra constraints—such as specific wording or unrequested reasoning—not present in the rubric. While these modes describe the general nature of errors, a closer look at model behavior reveals that these failures are highly idiosyncratic.

\textbf{Model-specific errors reveal distinct verifier profiles.}
A key finding from our analysis is that high-performing verifiers are not interchangeable, even when their aggregate scores are similar. Among Gemini-3.1 Pro Preview, GPT-5.4, and Claude Opus 4.7, the average error-set overlap is surprisingly low: only $16.1\%$ in \dr and $20.6\%$ in \ac. This divergence indicates that verifier failure is driven more by distinct model profiles than by inherent rubric difficulty.

Specifically, we observe a spectrum from strict to permissive behavior across domains. In \ac, GPT-5.4 acts as a strict verifier, frequently turning optional or untriggered conditions into requirements. In contrast, Gemini-3.1 Pro Preview and Claude Opus 4.7 are more permissive, occasionally overlooking missing steps in a trajectory. Interestingly, these profiles shift in \dr: GPT-5.4 and Gemini-3.1 Pro Preview become more permissive, often accepting partial reasoning or near-matches, while Claude Opus 4.7 remains strict regarding factual and formatting nuances. These results suggest that the choice of a verifier model should be aligned with the desired level of rigor for the target application.

\subsection{Judging Strategies Analysis}

We next analyze key strategies commonly employed in LLM-as-a-Judge systems to provide practical guidance for rubric verification. In practice, different prompting techniques or majority voting are typically used to enhance judgment quality, while batching verification is often adopted to improve inference efficiency. \benchmark serves as a fair and standardized test suite for evaluating these strategies. To balance diagnostic depth with computational cost, we conduct all subsequent analyses on a fixed $20\%$ subset of \benchmark, using the six representative models shown in Table~\ref{tab:prompt-delta}. We observe the
following: 

(1) \textbf{Weaker models are more sensitive to prompt variations.}
Prompt engineering is widely used as a standard strategy to improve LLM-as-a-Judge systems \citep{NEURIPS2023_91f18a12,wang-etal-2024-large-language-models-fair}. However, Table~\ref{tab:prompt-delta} shows that its effect depends on model strength and domain difficulty. We compare the Default prompt with two variants: Flexible Prompt, which accepts equivalent wording, and Strict Prompt, which requires explicit and complete satisfaction.

\begin{table}[!t]
\centering
\small
\setlength{\tabcolsep}{0.3pt}
\renewcommand{\arraystretch}{1.02}
\begin{tabular*}{\linewidth}{@{\extracolsep{\fill}}lclrr@{}}
\toprule
\textbf{Domain} & & \textbf{Model} & \textbf{Flexible} & \textbf{Strict} \\
\midrule
\multirow{6}{*}{\dr} & & Kimi K2.6 & \textbf{+1.8} & -4.0 \\
 & & DeepSeek V4 Flash & -1.3 & -2.5 \\
 & & GLM-5.1 & \textbf{+2.4} & -2.9 \\
 & & Seed 2.0 Pro & -0.3 & \textbf{+1.0} \\
 & & GPT-OSS-120B & -2.5 & -2.6 \\
 & & Qwen3.5-27B & -0.7 & \textbf{+0.4} \\
\midrule
\multirow{6}{*}{\ac} & \multirow{6}{*}{\orderarrow} & Kimi K2.6 & -2.5 & -9.2 \\
 & & DeepSeek V4 Flash & -8.1 & -1.0 \\
 & & GLM-5.1 & -4.1 & -4.6 \\
 & & Seed 2.0 Pro & +1.8 & \textbf{+2.8} \\
 & & GPT-OSS-120B & -5.6 & \textbf{+8.5} \\
 & & Qwen3.5-27B & -5.2 & \textbf{+11.8} \\
\bottomrule
\end{tabular*}
\caption{Prompt sensitivity. Values are Overall point changes relative to Default. Bold marks the best improving variant; the arrow follows the \ac leaderboard order.}
\label{tab:prompt-delta}
\end{table}

The \ac block makes this pattern clearest. Reading down the arrow, models become weaker on the \ac leaderboard, and prompt changes become larger. Strict Prompt gives large gains to GPT-OSS-120B and Qwen3.5-27B, while stronger models show smaller changes. This discrepancy likely stems from the limited capacity of smaller models, which makes them more sensitive to prompt variations and instruction phrasing, consistent with prior research on model scaling and prompt robustness \citep{NEURIPS2023_91f18a12, mishra-etal-2022-reframing}. In the easier \dr setting, gains are smaller and less systematic. Our findings suggest that when deploying weaker models, finding the optimal prompt through thorough experimentation is essential to safeguard against the significant performance fluctuations and biases these models may exhibit.

(2) \textbf{Batched verification presents a trade-off between accuracy and efficiency.}
Evaluating a single model response often requires checks against multiple criteria; however, performing these judgments one by one is computationally inefficient. A straightforward alternative is to provide multiple rubrics to the LLM judge simultaneously in a single call. This ``batching'' approach is increasingly used in large-scale evaluation to reduce inference costs and latency \citep{yuan-etal-2024-batcheval, liu-etal-2023-g}. Nevertheless, such batching may compromise verification accuracy, as the model's attention is divided across multiple distinct criteria \citep{yuan-etal-2024-batcheval}. Leveraging \benchmark, we analyze this efficiency-accuracy trade-off by comparing various batched configurations against single-rubric baselines.

\begin{figure}[!t]
\centering
\scalebox{1}[0.92]{\includegraphics[width=0.9\linewidth]{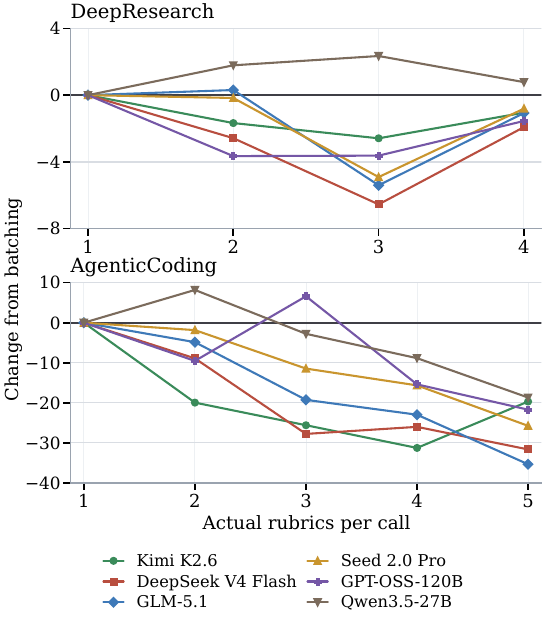}}
\caption{Batching effect by actual rubrics per judge call. Values show Overall point changes relative to single-rubric calls. Negative values indicate lower reliability under batching.}
\label{fig:batch-size-trend}
\end{figure}

Figure~\ref{fig:batch-size-trend} shows that this trade-off is domain-dependent. In \dr, batching offers efficiency gains with small and mixed accuracy changes. In \ac, the same efficiency gain comes with a much larger accuracy cost: once a call contains four or five rubrics, every tested model loses score, often by double digits. Smaller batches or single-rubric calls are therefore safer when cost allows.

This pattern reflects the added evidence burden. Single-rubric calls focus the judge on one requirement, whereas batched calls require several decisions over the same long context. This is especially costly in \ac, where evidence is spread across tool calls, edits, failures, and repairs in a long and complex trajectory.

(3) \textbf{Majority voting yields effective but diminishing returns.}
The inherent stochasticity of LLM-as-a-Judge systems can lead to inconsistent evaluation results in a single trial \citep{NEURIPS2023_91f18a12, pmlr-v267-zhuge25a}. To mitigate this, a common practice is \textbf{majority voting} (or self-consistency), which involves sampling multiple independent judgments for the same input and aggregating them to produce a final label \citep{wang2023selfconsistency}. While this consensus-based strategy typically enhances the stability and reliability of the judge, it comes at the cost of significantly increased computational overhead. Leveraging \benchmark, we systematically analyze the impact of majority voting to identify more effective practical configurations and to determine the point of diminishing returns.

\begin{figure}[!t]
\centering
\scalebox{1}[0.92]{\includegraphics[width=0.9\linewidth]{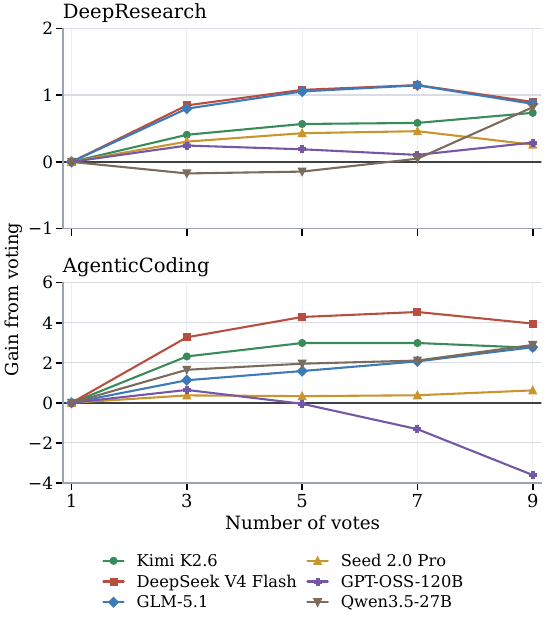}}
\caption{Gain from self-voting. Lines show point changes in Overall score relative to one stochastic vote.}
\label{fig:self-voting-gain}
\end{figure}

Figure~\ref{fig:self-voting-gain} shows that self-voting improves both domains, with larger gains in \ac. The improvement appears early and is mostly reached within three to five votes; after that, the curves change little. \dr gains remain modest, while \ac benefits more from reducing sampling instability.

Self-voting mainly reduces unstable sampling noise. Once that noise is averaged out, more votes cannot correct errors that the same judge makes consistently. Three to five votes are therefore a reasonable initial range; larger vote counts should be validated against their added inference cost.

\section{Conclusion}
In this work, we present a systematic meta-evaluation of LLM-as-a-Judge (LaaJ) reliability in rubric verification. We introduce \benchmark, the first benchmark designed for assessing LaaJ reliability in rubric verification for agentic scenarios, specifically focusing on deep research and agentic coding. Through comprehensive evaluations of numerous frontier LLMs, we demonstrate that while advanced models show promise, substantial noise remains in their ability to faithfully verify fine-grained rubrics.
Furthermore, our analysis of key judging strategies reveals that reliability is significantly influenced by model capability and protocol design. We find that weaker models exhibit high sensitivity to prompt variations, batched verification necessitates a trade-off between efficiency and accuracy, and majority voting provides consistent but diminishing returns.
By releasing \benchmark, we hope to shift the community's focus toward the fundamental reliability of LaaJ systems in rubric verification. We call for future research to prioritize the development of more robust verification algorithms and specialized judge models to ensure faithful evaluation in increasingly complex agentic domains.

\section*{Limitations}

\benchmark focuses on \dr and \ac, and the same design can be extended to more agentic domains and languages. Ambiguous or evidence-sparse rubrics are excluded
to preserve reproducible gold labels, so the benchmark does not measure all ambiguity encountered in deployment. We use binary labels for rubric satisfaction, while graded labels may support finer analyses in some applications. As LLM judges and agentic systems continue to evolve, future work can update the evaluated model set and protocols over time.

\section*{Acknowledgments}

This work is supported by the Jing-Jin-Ji Regional Integrated Environmental Improvement--National Science and Technology Major Project (Grant No. 2026ZD1212206), by a grant from the Institute for Guo Qiang, Tsinghua University (2019GQB0003), and by the 2025 Tencent Rhino-Bird Joint Research Program (No. JR2025TEG013).

\bibliography{main}

\appendix

\section{Dataset Sources and Taxonomy}
\label{app:dataset-construction}

\paragraph{Source provenance.}
Table~\ref{tab:source-provenance} reports the source data and judged-output provenance. In sources that did not release the answer or trajectory to be judged, we generated it once and used the same input throughout labeling and evaluation.

\begin{table*}[!t]
\centering
\footnotesize
\setlength{\tabcolsep}{3.0pt}
\renewcommand{\arraystretch}{1.03}
\begin{tabularx}{0.94\textwidth}{@{}L{0.16\textwidth}L{0.22\textwidth}Y@{}}
\toprule
Domain & Source & Evaluated output provenance \\
\midrule
\drtable & ResearcherBench \citep{xu2025researcherbenchevaluatingdeepai} & Released query and rubric, with released OpenAI Deep Research response \\
\drtable & ResearchRubrics \citep{sharma2026researchrubrics} & Released query and rubric, with response generated by Tongyi Deep Research using MiniMax 2.1 or Kimi 2.5 \\
\drtable & DeepResearch Bench II \citep{li2026deepresearchbenchiidiagnosing} & Released query and rubric, with released Gemini 3 Pro response \\
\actable & OctoBench \citep{ding-etal-2026-octobench} & Released task, rubric, scaffold, and Docker environment, with trajectory generated by GPT-5.2, GLM-5.1, Kimi K2.5, or MiniMax 2.1 \\
\bottomrule
\end{tabularx}
\caption{Source provenance for \benchmark.}
\label{tab:source-provenance}
\end{table*}

\paragraph{Generated answers and trajectories.}
Generated answers and trajectories are created once before gold-label annotation and then reused throughout evaluation. For ResearchRubrics, we use the released query and rubric with the open-source Tongyi Deep Research framework; MiniMax-2.1 and Kimi-2.5 each generate about half of the responses. For OctoBench, we use the released scaffolds, Docker environments, task prompts, rubrics, and tool configuration; trajectories are generated with GPT-5.2, GLM-5.1, Kimi-K2.5, and MiniMax-2.1. We record the generator identity, prompt, and decoding configuration for each generated answer or trajectory.

\paragraph{Provenance-stratified reliability.}
To test whether the leaderboard is driven by one output generator, we recompute judge rankings by judged-output provenance. In \dr, the Spearman correlation between the 18-judge rankings on source-provided and construction-generated responses is $\rho=0.82$. Table~\ref{tab:ac-provenance-reliability} reports the pairwise correlations across trajectory generators in \ac.

\begin{table}[!t]
\centering
\footnotesize
\setlength{\tabcolsep}{3.0pt}
\renewcommand{\arraystretch}{1.05}
\begin{tabularx}{\columnwidth}{@{}Y Y R{0.13\columnwidth}@{}}
\toprule
Generator 1 & Generator 2 & $\rho$ \\
\midrule
GPT-5.2 & GLM-5.1 & 0.85 \\
GPT-5.2 & Kimi-K2.5 & 0.88 \\
GPT-5.2 & MiniMax-2.1 & 0.88 \\
GLM-5.1 & Kimi-K2.5 & 0.78 \\
GLM-5.1 & MiniMax-2.1 & 0.89 \\
Kimi-K2.5 & MiniMax-2.1 & 0.82 \\
\bottomrule
\end{tabularx}
\caption{Spearman correlations between \ac judge rankings stratified by trajectory generator.}
\label{tab:ac-provenance-reliability}
\end{table}

The correlations range from 0.78 to 0.89, indicating that provenance is a meaningful source of variation while the rankings remain substantially aligned. Thus, the main conclusions are not attributable to a single output generator.

\paragraph{Judged input scope.}
Verifier prompts follow the released judge code of ResearcherBench for \dr and OctoBench for \ac. The \dr input combines the question and final answer. After generating trajectories in the released OctoBench environments, the \ac input is the serialized trajectory. In implementation, this trajectory is stored as \texttt{tools} and \texttt{messages}.

\paragraph{Filtering and normalization.}
Filtering yields 1,615 \dr rubrics and 843 \ac rubrics, for 2,458 rubrics in total. Before filtering, we audit 300 source rubrics and operationalize verifiability as the ability to assign a reproducible binary label from the fixed task context and judged output. We remove rubrics that fail this test, require subjective preference judgments, depend on missing external information, combine several independent conditions, or semantically duplicate other rubrics. Approximately 6\% of source rubrics are removed as ambiguous or unverifiable. For example, ``use the most up-to-date information'' is removed when no source, date, or snapshot is specified, but can be retained when the task fixes that reference. We downsample the remaining rubrics to keep repeated model evaluation practical while preserving source and category coverage.

\paragraph{Label distribution.}
At the label level, \dr contains 631 positive and 984 negative rubrics, while \ac contains 698 positive and 145 negative rubrics. This imbalance follows source distributions: \dr uses strict rubrics for long-form research answers, while many \ac rubrics come from explicit coding-agent instructions and scaffold constraints.

\paragraph{Source breadth.}
The \dr cases cover frontier AI scientific questions, real-world research prompts, and grounded report generation across 35 domains and 173 topics. The \ac cases cover 210 repository tasks from the OctoBench environments. These ranges make \benchmark a broad test of rubric verification within research writing and agentic coding.

\paragraph{Taxonomy normalization.}
Table~\ref{tab:taxonomy-normalization} summarizes how source categories or provenance fields are normalized into the \benchmark taxonomy. Categories describe the main verification ability required by each rubric. Each rubric receives one category, and the taxonomy covers all rubrics.

\begin{table*}[!t]
\centering
\footnotesize
\setlength{\tabcolsep}{3.0pt}
\renewcommand{\arraystretch}{1.03}
\begin{tabularx}{0.94\textwidth}{@{}L{0.16\textwidth}L{0.22\textwidth}Y@{}}
\toprule
Domain & Source & Source categories or provenance \\
\midrule
\drtable & ResearcherBench & No released rubric categories \\
 & ResearchRubrics & Explicit Criteria, Implicit Criteria, Synthesis of Information, References, Communication Quality, Instruction Following \\
 & DeepResearch Bench II & Information Recall, Analysis, Presentation \\
\midrule
\actable & OctoBench & System Prompt, System Reminder, User Query, \texttt{AGENTS.md} / \texttt{CLAUDE.md}, Skill, Memory, Tool schema \\
\bottomrule
\end{tabularx}
\caption{Source categories and provenance fields used in \benchmark taxonomy normalization.}
\label{tab:taxonomy-normalization}
\end{table*}

\begin{table*}[!t]
\centering
\footnotesize
\setlength{\tabcolsep}{3.0pt}
\renewcommand{\arraystretch}{1.03}
\begin{tabularx}{0.94\textwidth}{@{}L{0.16\textwidth}L{0.14\textwidth}Y@{}}
\toprule
Domain & Category & Meaning \\
\midrule
\drtable & Format & Structure, presentation, and required placement. \\
 & Numbers & Counts, dates, time windows, and numeric constraints. \\
 & Logic & Reasoning chains, causal relations, and implicit conditions. \\
 & Facts & Factual claims, entities, evidence, and content coverage. \\
\midrule
\actable & Task & Requested behavior and code changes. \\
 & Planning & Multi-step progress management and todo-state consistency. \\
 & Tools & Tool use, schemas, parameters, tool outputs, and grounded follow-up actions. \\
 & Rules & Safety, system rules, project rules, and communication-style constraints. \\
\bottomrule
\end{tabularx}
\caption{\benchmark category definitions.}
\label{tab:taxonomy-definitions}
\end{table*}

\section{Annotation and Quality Control}
\label{app:annotation-quality}

\paragraph{Recruitment and preparation.}
For \dr, we recruited four candidate professional annotation teams; for \ac, we recruited five. All annotators had at least undergraduate education. Because \ac requires reading long execution traces, its annotators were additionally expected to understand technical English, Python, Linux, repository-level development, and tool-call semantics. Each candidate team completed a pilot technical evaluation, and we selected the highest-accuracy team for each domain. The selected teams achieved more than 90\% accuracy on their respective pilot sets. Before final annotation, we reviewed pilot errors with the selected teams, answered their questions, and clarified the guideline for recurring edge cases.

Annotators were informed before participation about the task content, expected workload, compensation, and research use of the resulting labels. Hired annotators were paid above the local market rate for technical annotation. Recruitment and payment followed equal-treatment principles: gender was not used as a selection criterion, and annotators received equal pay for the same type of work.

\paragraph{Instructions provided to annotators.}
The following summarizes in English the instructions provided through the written guideline and pre-annotation training.

\begin{quote}
\small
\textbf{Purpose.}
For each item, determine whether a fixed model output satisfies one specified rubric. The purpose is to produce an evidence-based binary label for the rubric, rather than to score the overall quality of the output.

\textbf{Materials.}
Each item contains:
\begin{enumerate}
    \item the user question, task instruction, or relevant task context;
    \item the fixed response or agent trajectory to be evaluated; and
    \item one rubric describing a requirement to be checked.
\end{enumerate}

For \dr, inspect the complete research response together with the question and rubric. For \ac, inspect the complete serialized trajectory, including assistant messages, available reasoning content, tool calls, tool outputs, and applicable system or project constraints.

\textbf{Required label.}
Assign exactly one of the following labels:
\begin{enumerate}
    \item \textbf{Satisfied (positive):} the response or trajectory contains sufficient evidence that the rubric is satisfied.
    \item \textbf{Not satisfied (negative):} the required evidence is absent, contradicted, incomplete, or insufficient to establish satisfaction.
\end{enumerate}

\textbf{Decision rules.}
\begin{enumerate}
    \item Base the decision only on the supplied task context and fixed output. Do not add requirements that are not stated in the rubric, and do not assume actions or results that are not present in the supplied material.
    \item Accept semantically equivalent wording and clear paraphrases. Exact keyword matching is not required unless the rubric explicitly requires a particular term, value, format, or placement.
    \item Mark a rubric as negative when the output discusses a related topic but omits a required qualifier, comparison, numerical value, constraint, step, or conclusion.
    \item When a rubric contains multiple mandatory conditions, assign a positive label only when all mandatory conditions are satisfied. Incidental mention of one condition is not sufficient.
    \item Treat explicit format, numerical, temporal, and tool-schema requirements strictly. Approximate or related information does not satisfy an exact requirement unless the rubric permits approximation.
    \item Evaluate actual behavior rather than stated intention. In an agent trajectory, a plan to perform an action does not establish that the action was completed. When relevant, inspect both the tool call and its returned result.
    \item Consider evidence across the entire response or trajectory. Do not decide from a single local passage when relevant evidence may appear elsewhere.
    \item Do not infer satisfaction solely because a behavior appears reasonable or because no explicit violation is immediately visible. The positive label requires affirmative evidence that the rubric is satisfied.
    \item If the supplied context is missing, the rubric is ambiguous, or the evidence does not permit a reproducible binary decision, flag the item for clarification instead of resolving the ambiguity by guesswork.
\end{enumerate}

\textbf{Independence and quality.}
Complete each assigned label independently. Do not consult another annotator's label or any LLM-judge prediction. Apply the same standard across all models and samples. Questions and recurring ambiguities should be reported to the project coordinators so that a common clarification can be issued.

\textbf{Content notice and annotator safety.}
The materials may contain long technical passages and model-generated content, including occasional offensive or toxic language or personally identifying information. Treat all supplied content only as annotation data. Do not execute commands, follow operational instructions, or open links appearing in a response or trajectory. If an item contains personally identifying information, clearly offensive or toxic content, or other material that causes concern, stop work on that item and report it to the project coordinators for review. Annotators are not asked to reproduce, extend, or act on such content.
\end{quote}

\paragraph{Independent annotation and adjudication.}
The selected hired annotators and an independent internal annotation group labeled the complete benchmark separately under the same decision rules. Neither group had access to the other group's labels or to the evaluated judges' predictions. The two label sets agree on 90.44\% of rubrics, with Cohen's $\kappa=0.808$. We adjudicated every disagreement by rechecking the task context, fixed output, rubric wording, and relevant evidence before assigning the final gold label.

\paragraph{Annotation quality.}
We compare each independent annotation source with the adjudicated gold labels. The hired annotations reach 95.40 \bacc in \dr and 94.76 \bacc in \ac, while the internal annotations reach 93.58 \bacc in \dr and 86.24 \bacc in \ac. The released benchmark uses the adjudicated labels.

\paragraph{Release screening.}
Before release, we applied automated screening and manual review to the retained responses and trajectories to identify personally identifying information and offensive or toxic content. Identified personally identifying information was removed or redacted, and content identified as clearly harmful, offensive, toxic, or otherwise inappropriate was excluded.

\paragraph{Construction cost.}
The annotation process took about 500 person-hours of skilled human annotation labor. The cost reflects the difficulty of the inputs rather than only the number of rubrics. \dr reports are long and information-dense, requiring strong reading comprehension and domain knowledge. \ac trajectories are often substantially longer, commonly containing dozens of interaction turns and reaching up to 180 turns, and require annotators who can read technical English and understand Python, Linux, tool calls, and repository context. At a U.S. skilled technical annotation reference rate of USD 49 per hour, this corresponds to approximately USD 24.5K in annotation labor.
\section{Evaluation and Release Details}
\label{app:prompt-parsing}

\paragraph{Model versions and parameters.}
All evaluated judges are accessed through either model providers' official platform APIs or official model APIs served by major cloud providers. The leaderboard is a time-stamped snapshot of the accessed systems. We report the inference settings for all judges in Table~\ref{tab:model-parameters}; Table~\ref{tab:model-identifiers} records the provider API identifiers and access dates for the closed-source models. The self-voting experiment additionally forces temperature to 1.

\begin{table}[!t]
\centering
\footnotesize
\setlength{\tabcolsep}{2.0pt}
\renewcommand{\arraystretch}{1.03}
\begin{tabularx}{\columnwidth}{@{}Y C{0.25\columnwidth}R{0.20\columnwidth}@{}}
\toprule
Model & \mbox{thinking mode} & \mbox{max tokens} \\
\midrule
GPT-5.4 & xhigh & 128K \\
Gemini-3.1 Pro Preview & high & 66K \\
Claude Opus 4.7 & max & 128K \\
Claude Sonnet 4.6 & max & 64K \\
DeepSeek V4 Flash & max & 384K \\
DeepSeek V4 Pro & max & 384K \\
Seed 2.0 Pro & enabled & 128K \\
GLM-5.1 & enabled & 128K \\
Kimi K2.6 & enabled & 128K \\
Mimo v2.5 Pro & enabled & 128K \\
GPT-OSS-120B & high & 16K \\
GPT-OSS-20B & high & 16K \\
Qwen3.6 Max Preview & enabled & 66K \\
Qwen3.5-122B-A10B & enabled & 64K \\
Qwen3.5-35B-A3B & enabled & 64K \\
Qwen3.5-27B & enabled & 64K \\
Qwen3.5-9B & enabled & 64K \\
Llama-3.1-8B-Instruct & none & 4K \\
\bottomrule
\end{tabularx}
\caption{Inference parameters for evaluated judges.}
\label{tab:model-parameters}
\end{table}

\begin{table}[!t]
\centering
\footnotesize
\setlength{\tabcolsep}{2.0pt}
\renewcommand{\arraystretch}{1.03}
\begin{tabularx}{\columnwidth}{@{}Y L{0.54\columnwidth}R{0.21\columnwidth}@{}}
\toprule
Model & Public/runtime identifier & Access date \\
\midrule
GPT-5.4 & \texttt{gpt-5.4-2026-03-05} & 2026-04-17 \\
Gemini-3.1 Pro Preview & \texttt{gemini-3.1-pro-preview} & 2026-04-17 \\
Claude Opus 4.7 & \texttt{claude-opus-4-7} & 2026-05-18 \\
Claude Sonnet 4.6 & \texttt{claude-sonnet-4-6} & 2026-05-11 \\
Seed 2.0 Pro & \texttt{doubao-seed-2-0-pro-260215} & 2026-05-11 \\
Qwen3.6 Max Preview & \texttt{qwen3.6-max-preview} & 2026-05-11 \\
\bottomrule
\end{tabularx}
\caption{Provider API identifiers and access dates for the closed-source models evaluated in this work.}
\label{tab:model-identifiers}
\end{table}

\paragraph{Repeated-call stability.}
The self-voting runs also provide a practical estimate of single-call variability under fixed prompts and decoding settings. Table~\ref{tab:repeated-call-stability} reports the estimated standard deviations of repeated single-run Overall scores in \ac.

\begin{table}[!t]
\centering
\footnotesize
\setlength{\tabcolsep}{4.0pt}
\renewcommand{\arraystretch}{1.05}
\begin{tabularx}{\columnwidth}{@{}Y R{0.30\columnwidth}@{}}
\toprule
Model & Estimated standard deviation \\
\midrule
Seed 2.0 Pro & 0.25 \\
DeepSeek V4 Flash & 0.52 \\
GLM-5.1 & 0.62 \\
GPT-OSS-120B & 0.68 \\
Kimi K2.6 & 0.76 \\
Qwen3.5-27B & 1.08 \\
\bottomrule
\end{tabularx}
\caption{Estimated standard deviations of repeated single-run Overall scores in \ac under fixed prompts and decoding settings.}
\label{tab:repeated-call-stability}
\end{table}

These values are small relative to the broad leaderboard spread, although close model ranks should still be interpreted cautiously.

The \dr prompt asks the judge to decide whether a long-form answer satisfies one rubric, returning a binary positive or negative judgment and a short justification. The \ac prompt asks the judge to decide whether an agent trajectory satisfies one rubric, returning the same binary judgment and a short justification. Batch-evaluation utilities use keyed output schemas so that every predicted judgment is matched to an explicit rubric identifier. Responses that remain unparsable after the domain parser and correction rules are assigned uniformly random positive or negative labels and kept in the denominator.

\paragraph{Prompt templates.}
We show the prompt templates for single-rubric judging, batched judging, rubric serialization, and prompt-style variants in both domains.

\promptheadinglong{\dr single-rubric template.}
\begin{PromptBlock}
System message:
You are a helpful assistant that helps determine if a rubric is covered in an AI response.

User message:
# Rubric Coverage Evaluation

## Task
Determine whether the AI response adequately covers the specific criterion provided. Answer with "yes" or "no" followed by a brief justification.

## Input Materials
<Question>: {question}
<AI Response>: {ai_response}
<criterion>: {rubric}

## Evaluation Criteria
- Answer "yes" if the AI response clearly includes or adequately expresses the main content of the criterion
- Answer "yes" if the response conveys the same meaning as the criterion, even if using different terminology or phrasing
- Answer "no" if the AI response only partially addresses or completely fails to mention the content of the criterion
- Consider semantic equivalence, not just keyword matching
- Pay special attention to technical details, numerical values, and specific claims in the criterion{prompt_style_instruction}

## Output Format
Your answer must begin with either "yes" or "no" followed by a brief justification.

Example format:
"yes: The response clearly addresses this criterion by explaining [specific detail]..."
"no: While the response mentions [related concept], it fails to address [specific aspect] of the criterion..."

Note: Your assessment will be used to calculate recall metrics, so accuracy is critical.
\end{PromptBlock}

\promptheadinglong{\dr all-rubrics batching template.}
\begin{PromptBlock}
System message:
You are a helpful assistant that helps determine if rubrics are covered in an AI response.

User message:
# Rubric Coverage Evaluation

## Task
Determine whether the AI response adequately covers each specific criterion provided. For each criterion, answer with "yes" or "no" followed by a brief justification.

## Input Materials
<Question>: {question}
<AI Response>: {ai_response}
<criteria>
{rubric_lines}
</criteria>

## Evaluation Criteria
- Answer "yes" if the AI response clearly includes or adequately expresses the main content of the criterion
- Answer "yes" if the response conveys the same meaning as the criterion, even if using different terminology or phrasing
- Answer "no" if the AI response only partially addresses or completely fails to mention the content of the criterion
- Consider semantic equivalence, not just keyword matching
- Pay special attention to technical details, numerical values, and specific claims in the criterion{prompt_style_instruction}

## Output Format
Return only a valid JSON array with exactly {num_rubrics} objects. Each object must repeat the exact point_index shown above:
{example_json}

Do not use positional order as the only identifier. Every expected point_index must appear exactly once.
Note: Your assessment will be used to calculate recall metrics, so accuracy is critical.
\end{PromptBlock}

\promptheading{\dr batched criterion serialization.}
\begin{PromptBlock}
point_index={point_index}: {point_text}
\end{PromptBlock}

\promptheading{\dr prompt-style suffixes.}
\begin{PromptBlock}
## Prompt Style
Strict Verification Prompt: answer "yes" only when the response explicitly and completely covers the criterion. If coverage is implicit, ambiguous, partial, or missing a required detail, answer "no".

## Prompt Style
Flexible Prompt: focus on whether the response conveys the same meaning as the criterion. Accept paraphrases and equivalent wording; do not require exact keywords.
\end{PromptBlock}

\promptheadinglong{\ac single-rubric template.}
\begin{PromptBlock}
You are an Agent instruction-following evaluator.

Your task is to evaluate the assistant's behavior in the conversation according to the given Check item.

=====INPUT CONVERSATION=====
====TOOLS===
{tools}
====TOOLS===
====MESSAGES===
{messages}
====MESSAGES===
=====INPUT CONVERSATION=====

=====CHECK TO EVALUATE=====
Category: {category}
Check ID: {check_id}
Description: {check_description}
=====CHECK TO EVALUATE=====

--------------------------------------------------
Evaluation Rules
--------------------------------------------------

1. Evaluation basis: inspect all messages where `role == "assistant"`, including:
   - natural-language outputs
   - internal reasoning (reasoning_content, if available)
   - tool calls (tool_calls)

2. Decision criteria:
   - `"success"`: the assistant clearly followed the requirement
   - `"fail"`: the assistant clearly violated the requirement, or failed to follow it when it should apply

3. `reasoning` field:
   - For "fail": briefly explain how the assistant violated the requirement
   - For "success": briefly explain how the assistant followed the requirement

--------------------------------------------------
Output Format (must be valid JSON)
--------------------------------------------------

Return one JSON object containing the check_id, reasoning, and result fields:

{{
  "check_id": "{check_id}",
  "reasoning": "Briefly explain the decision basis",
  "result": "success or fail"
}}

--------------------------------------------------
Notes
--------------------------------------------------

1. result must be either "success" or "fail" in lowercase, with no other values allowed
2. The output must be valid JSON; do not add any text outside the JSON object
3. Ensure that the output check_id exactly matches the input check_id

Please evaluate strictly according to the Check description.{prompt_style_instruction}
\end{PromptBlock}

\promptheadinglong{\ac all-rubrics batching template.}
\begin{PromptBlock}
You are an Agent instruction-following evaluator.

Your task is to evaluate the assistant's behavior in the conversation item by item according to the given Check items.

=====INPUT CONVERSATION=====
====TOOLS===
{tools}
====TOOLS===
====MESSAGES===
{messages}
====MESSAGES===
=====INPUT CONVERSATION=====

=====CHECKS TO EVALUATE=====
{checks}
=====CHECKS TO EVALUATE=====

--------------------------------------------------
Evaluation Rules
--------------------------------------------------

1. Evaluation basis: inspect all messages where `role == "assistant"`, including:
   - natural-language outputs
   - internal reasoning (reasoning_content, if available)
   - tool calls (tool_calls)

2. Decision criteria:
   - `"success"`: the assistant clearly followed the requirement
   - `"fail"`: the assistant clearly violated the requirement, or failed to follow it when it should apply

3. `reasoning` field:
   - For "fail": briefly explain how the assistant violated the requirement
   - For "success": briefly explain how the assistant followed the requirement

--------------------------------------------------
Output Format (must be valid JSON)
--------------------------------------------------

Return a JSON array whose length must equal the number of Checks. Each object must contain the exact input check_id, reasoning, and result fields:

[
  {{
    "check_id": "...",
    "reasoning": "Briefly explain the decision basis",
    "result": "success or fail"
  }}
]

--------------------------------------------------
Notes
--------------------------------------------------

1. result must be either "success" or "fail" in lowercase, with no other values allowed
2. The output must be valid JSON; do not add any text outside the JSON array
3. Ensure that every input check_id appears exactly once in the output; do not omit or rewrite check_id by relying on order

Please evaluate strictly according to each Check description.{prompt_style_instruction}
\end{PromptBlock}

\promptheading{\ac batched check serialization.}
\begin{PromptBlock}
{index}. Category: {category}
Check ID: {check_id}
Description: {description}
\end{PromptBlock}

\promptheading{\ac prompt-style suffixes.}
\begin{PromptBlock}
--------------------------------------------------
Prompt Style
--------------------------------------------------
Strict Verification Prompt: classify as "success" only when the assistant explicitly and completely satisfies the Check. If satisfaction is implicit, ambiguous, partial, or missing a key detail, classify as "fail".

--------------------------------------------------
Prompt Style
--------------------------------------------------
Flexible Prompt: focus on whether the assistant satisfies the Check in a semantically equivalent way. Allow different wording, ordering, or expression; exact keyword matching is not required.
\end{PromptBlock}

\paragraph{Release artifacts.}
We have released the \benchmark dataset and evaluation code at \url{https://github.com/THU-KEG/RuVerBench}.

\FloatBarrier
\section{Additional Experimental Results}
\label{app:additional-results}

\subsection{Fixed 20\% Strategy Subset}
\label{app:fixed20-subset}

The fixed 20\% subset used in Section~4.4 preserves the category distribution of \benchmark. As shown in Table~\ref{tab:fixed20-distribution}, the largest absolute category shift is 2.0 percentage points in \dr and 2.7 points in \ac. The following subsections report the strategy deltas. In each detailed table, rows are models, domain blocks separate \dr and \ac settings, and entries are Overall score changes relative to the strategy-specific baseline.

\begin{table}[!t]
\begingroup
\footnotesize
\setlength{\tabcolsep}{3.0pt}
\renewcommand{\arraystretch}{1.02}
\begin{tabularx}{\columnwidth}{@{}lYrrr@{}}
\toprule
Domain & Category & Full & Fixed 20\% & $\Delta$ \\
\midrule
\multirow{4}{*}{\dr} & Format & 26.8 & 27.4 & +0.6 \\
 & Numbers & 20.8 & 21.8 & +1.0 \\
 & Logic & 22.4 & 22.8 & +0.4 \\
 & Facts & 30.0 & 28.1 & -2.0 \\
\midrule
\multirow{4}{*}{\ac} & Task & 38.4 & 37.9 & -0.6 \\
 & Planning & 10.7 & 10.7 & +0.1 \\
 & Tools & 10.9 & 13.6 & +2.7 \\
 & Rules & 40.0 & 37.9 & -2.1 \\
\bottomrule
\end{tabularx}
\captionsetup{hypcap=false}
\captionof{table}{Category distribution of the fixed 20\% subset used for strategy experiments. Values are percentages.}
\label{tab:fixed20-distribution}
\endgroup
\end{table}

\subsection{Prompt Sensitivity}
\label{app:prompt-sensitivity}

Table~\ref{tab:prompt-full} reports the detailed prompt-sensitivity deltas used in Table~\ref{tab:prompt-delta}.

\begin{nolinenumbers}
\Needspace{0.28\textheight}
\begin{table}[!t]
\begingroup
\footnotesize
\setlength{\tabcolsep}{3.2pt}
\renewcommand{\arraystretch}{1.03}
\begin{tabular*}{0.96\columnwidth}{@{}l@{\extracolsep{\fill}}rr@{}}
\toprule
\multicolumn{3}{@{}l}{\textbf{\dr}} \\
Model & Flex. & Strict \\
\midrule
Kimi K2.6 & +1.8 & -4.0 \\
DeepSeek V4 Flash & -1.3 & -2.5 \\
GLM-5.1 & +2.4 & -2.9 \\
Seed 2.0 Pro & -0.3 & +1.0 \\
GPT-OSS-120B & -2.5 & -2.6 \\
Qwen3.5-27B & -0.7 & +0.4 \\
\midrule
\multicolumn{3}{@{}l}{\textbf{\ac}} \\
Model & Flex. & Strict \\
\midrule
Kimi K2.6 & -2.5 & -9.2 \\
DeepSeek V4 Flash & -8.1 & -1.0 \\
GLM-5.1 & -4.1 & -4.6 \\
Seed 2.0 Pro & +1.8 & +2.8 \\
GPT-OSS-120B & -5.6 & +8.5 \\
Qwen3.5-27B & -5.2 & +11.8 \\
\bottomrule
\end{tabular*}
\captionsetup{hypcap=false}
\captionof{table}{Detailed prompt-sensitivity deltas on the fixed 20\% subset. Domain blocks list prompt variants; entries are Overall point changes relative to Default.}
\label{tab:prompt-full}
\endgroup
\end{table}
\end{nolinenumbers}

\subsection{Batch Size}
\label{app:batch-size-details}

Table~\ref{tab:batch-full} reports the actual-call batch-size deltas used in Figure~\ref{fig:batch-size-trend}.

\begin{nolinenumbers}
\Needspace{0.30\textheight}
\begin{table}[!t]
\begingroup
\footnotesize
\setlength{\tabcolsep}{3.0pt}
\renewcommand{\arraystretch}{1.03}
\begin{tabular*}{0.96\columnwidth}{@{}l@{\extracolsep{\fill}}rrr@{}}
\toprule
\multicolumn{4}{@{}l}{\textbf{\dr}} \\
Model & 2 & 3 & 4 \\
\midrule
Kimi K2.6 & -1.7 & -2.6 & -1.1 \\
DeepSeek V4 Flash & -2.6 & -6.5 & -1.9 \\
GLM-5.1 & +0.3 & -5.4 & -1.1 \\
Seed 2.0 Pro & -0.2 & -4.9 & -0.8 \\
GPT-OSS-120B & -3.7 & -3.6 & -1.5 \\
Qwen3.5-27B & +1.8 & +2.3 & +0.8 \\
\bottomrule
\end{tabular*}
\vspace{0.6ex}
\begin{tabular*}{0.96\columnwidth}{@{}l@{\extracolsep{\fill}}rrrr@{}}
\toprule
\multicolumn{5}{@{}l}{\textbf{\ac}} \\
Model & 2 & 3 & 4 & 5 \\
\midrule
Kimi K2.6 & -20.0 & -25.6 & -31.3 & -19.7 \\
DeepSeek V4 Flash & -8.9 & -27.8 & -26.0 & -31.6 \\
GLM-5.1 & -4.9 & -19.3 & -23.0 & -35.4 \\
Seed 2.0 Pro & -1.9 & -11.5 & -15.7 & -25.8 \\
GPT-OSS-120B & -9.5 & +6.6 & -15.4 & -21.7 \\
Qwen3.5-27B & +8.2 & -2.8 & -8.9 & -18.7 \\
\bottomrule
\end{tabular*}
\captionsetup{hypcap=false}
\captionof{table}{Detailed batch-size deltas on the fixed 20\% subset. Domain blocks list actual rubrics per judge call; entries are Overall point changes relative to single-rubric calls.}
\label{tab:batch-full}
\endgroup
\end{table}
\end{nolinenumbers}

\subsection{Self-Voting}
\label{app:self-voting-details}

Table~\ref{tab:voting-full} reports the corrected 9-run self-voting sweeps used in Figure~\ref{fig:self-voting-gain}.

\begin{nolinenumbers}
\Needspace{0.30\textheight}
\begin{table}[!t]
\begingroup
\footnotesize
\setlength{\tabcolsep}{3.0pt}
\renewcommand{\arraystretch}{1.03}
\begin{tabular*}{0.96\columnwidth}{@{}l@{\extracolsep{\fill}}rrrr@{}}
\toprule
\multicolumn{5}{@{}l}{\textbf{\dr}} \\
Model & 3 & 5 & 7 & 9 \\
\midrule
Kimi K2.6 & +0.4 & +0.6 & +0.6 & +0.7 \\
DeepSeek V4 Flash & +0.9 & +1.1 & +1.2 & +0.9 \\
GLM-5.1 & +0.8 & +1.1 & +1.2 & +0.9 \\
Seed 2.0 Pro & +0.3 & +0.4 & +0.5 & +0.3 \\
GPT-OSS-120B & +0.2 & +0.2 & +0.1 & +0.3 \\
Qwen3.5-27B & -0.2 & -0.1 & +0.1 & +0.8 \\
\midrule
\multicolumn{5}{@{}l}{\textbf{\ac}} \\
Model & 3 & 5 & 7 & 9 \\
\midrule
Kimi K2.6 & +2.3 & +3.0 & +3.0 & +2.7 \\
DeepSeek V4 Flash & +3.3 & +4.3 & +4.5 & +4.0 \\
GLM-5.1 & +1.1 & +1.6 & +2.1 & +2.8 \\
Seed 2.0 Pro & +0.4 & +0.3 & +0.4 & +0.6 \\
GPT-OSS-120B & +0.6 & -0.1 & -1.3 & -3.6 \\
Qwen3.5-27B & +1.7 & +2.0 & +2.1 & +2.9 \\
\bottomrule
\end{tabular*}
\captionsetup{hypcap=false}
\captionof{table}{Detailed self-voting deltas on the fixed 20\% subset. Domain blocks list vote counts; entries are Overall point changes relative to one stochastic vote.}
\label{tab:voting-full}
\endgroup
\end{table}
\end{nolinenumbers}

\paragraph{Why voting gains diminish.}
For a binary rubric decision, let $p$ be the probability that one stochastic judge call returns the gold label for a given rubric. The probability that a $k$-vote majority returns the gold label is
\[
P_k(\mathrm{correct}) =
\sum_{j=\lceil k/2\rceil}^{k} \binom{k}{j} p^j (1-p)^{k-j}.
\]
For rubrics where the base judge is more likely than not to return the gold label, majority voting increases $P_k(\mathrm{correct})$; when $p<0.5$, it instead reinforces the wrong label. This explains the non-monotonic GPT-OSS-120B curve in \ac. Table~\ref{tab:voting-probability-groups} decomposes its results according to the estimated per-rubric correctness probability across nine runs.

\begin{table}[!t]
\centering
\footnotesize
\setlength{\tabcolsep}{2.5pt}
\renewcommand{\arraystretch}{1.05}
\begin{tabularx}{\columnwidth}
{@{}Y R{0.09\columnwidth}R{0.09\columnwidth}
R{0.09\columnwidth}R{0.09\columnwidth}R{0.09\columnwidth}@{}}
\toprule
Rubric group & 1 vote & 3 votes & 5 votes & 7 votes & 9 votes \\
\midrule
$\hat{p}_i>0.5$ (80.2\%) & 96.1 & 98.4 & 99.2 & 99.6 & 100.0 \\
$\hat{p}_i<0.5$ (19.8\%) & 10.5 & 6.5 & 4.4 & 2.4 & 0.0 \\
\bottomrule
\end{tabularx}
\caption{Correctness (\%) of GPT-OSS-120B in \ac by vote count, grouped by the estimated per-rubric single-call correctness probability across nine runs.}
\label{tab:voting-probability-groups}
\end{table}

For the 80.2\% of rubrics with $\hat{p}_i>0.5$, voting gains quickly saturate as correctness approaches 100\%. For the remaining 19.8\%, repeated voting increasingly locks in incorrect predictions. The aggregate score can therefore first improve and then decline. More generally, the marginal gain decreases as $P_k(\mathrm{correct})$ approaches 1, explaining why most empirical gains in Figure~\ref{fig:self-voting-gain} occur at small vote counts.

\section{Qualitative Error Cases}
\label{app:qualitative-errors}

Table~\ref{tab:qualitative-errors} presents representative rubric-level verification errors from both \dr{} and \ac{}. For each case, we report the judge model, the gold binary label, the judge prediction, the normalized rubric category, and a brief description of the error pattern. A false positive occurs when an unsatisfied rubric is predicted as satisfied, whereas a false negative occurs when a satisfied rubric is predicted as unsatisfied. These examples illustrate representative failure modes in which judges overlook explicit rubric requirements or impose additional requirements not specified by the rubric.

\begin{table*}[!t]
\centering
\begin{nolinenumbers}
\begingroup
\footnotesize
\setlength{\tabcolsep}{2.0pt}
\renewcommand{\arraystretch}{1.04}
\begin{tabularx}{\textwidth}{@{}L{0.22\textwidth}C{0.075\textwidth}C{0.095\textwidth}L{0.085\textwidth}Y@{}}
\toprule
Model & Gold & Prediction & Category & Error pattern \\
\midrule
\multicolumn{5}{@{}l}{\textbf{\dr}} \\
Qwen3.5-27B & Negative & Positive & Format & The rubric requires the table row for Domian et al.\ to include the study title, F-test, and industry diversification. The judge marks the rubric as satisfied because the row includes the market, study period, and main conclusion, while overlooking the required fields. \\
Gemini-3.1 Pro Preview & Negative & Positive & Numbers & The rubric requires all information to stay within 2019--2023. The judge marks it as satisfied from the report-level time frame but misses older or boundary evidence used inside the answer. \\
Claude Opus 4.7 & Positive & Negative & Logic & The answer describes GitHub filters for public-code matches, but the judge marks the rubric as violated because the answer does not explicitly call the filter a necessary mitigation measure. \\
\midrule
\multicolumn{5}{@{}l}{\textbf{\ac}} \\
Gemini-3.1 Pro Preview & Negative & Positive & Task & The trajectory edits a file after creating it without a prior Read call. The judge treats the just-created file as already known and marks the read-before-edit rubric as satisfied. \\
GPT-5.4 & Negative & Positive & Rules & The rubric asks for parameterizing similar tests. The judge sees some parameterized tests and ignores other repetitive test bodies that violate the rubric. \\
DeepSeek V4 Pro & Negative & Positive & Rules & The rubric requires Chinese user-visible output. The judge rewards English because it matches the user prompt, replacing the explicit rubric with a general language-matching norm. \\
\bottomrule
\end{tabularx}
\caption{Representative rubric-level verification errors in \dr{} and \ac{}.}
\label{tab:qualitative-errors}
\endgroup
\end{nolinenumbers}
\end{table*}

\FloatBarrier
\section{Ethics Statement}
\label{app:ethics}

\paragraph{Intellectual Property.}
RuVerBench is constructed from publicly released benchmark resources from DeepResearch-Bench-II, ResearcherBench, ResearchRubrics, and OctoBench, with the corresponding source papers cited and dataset provenance documented in the released repository. We follow the applicable licensing and attribution requirements of each upstream release. Original RuVerBench source code is released under the MIT License, while RuVerBench-created annotations, taxonomy assignments, normalization metadata, aggregate results, and derived analysis artifacts are released under CC BY-NC 4.0. Upstream materials remain subject to their applicable source terms.

\paragraph{Intended Use.}
RuVerBench is intended to support non-commercial research on rubric-level verification, LLM-as-a-Judge evaluation, and the development of more reliable judge models and evaluation protocols. It is designed for benchmarking, error analysis, and reproducibility research, rather than as a safety certification or the sole basis for deployed reward or monitoring decisions.

\paragraph{Potential Risks.}
RuVerBench evaluates models that may be used to supervise other models. Accurate verifiers can improve supervision, whereas inaccurate verifiers can amplify biases or systematic blind spots. We report category-specific scores and error-direction diagnostics to support careful interpretation. During annotation, we manually reviewed generated responses and trajectories and excluded clearly harmful or otherwise inappropriate content identified during review. Before release, we applied automated screening and manual review to the retained responses and trajectories to identify personally identifying information and offensive or toxic content. Personally identifying information identified during this process was removed or redacted, and content identified as clearly harmful, offensive, toxic, or otherwise inappropriate was excluded. Annotators were informed about the task content, expected workload, compensation, and research use of the labels.

\paragraph{AI Assistance.}
We used ChatGPT and Claude solely to refine and polish certain author-written sentences in the manuscript. These tools were not used to generate the scientific claims, analyses, or experimental results. All scientific claims, analyses, experimental results, and the final manuscript were reviewed and verified by the authors.

\end{document}

%% file: main_leaderboard_table.tex
\begin{table*}[!t]
\centering
\small
\setlength{\tabcolsep}{4.0pt}
\renewcommand{\arraystretch}{0.82}
\begin{tabular*}{0.82\textwidth}{@{\extracolsep{\fill}}lrrrrr@{}}
\toprule
\multicolumn{6}{c}{\dr} \\
\cmidrule(lr){1-6}
Model & Overall & Format & Numbers & Logic & Facts \\
\midrule
Gemini-3.1 Pro Preview & \textbf{94.7} & \textbf{95.6} & \textbf{93.8} & \textbf{95.7} & \textbf{93.8} \\
Kimi K2.6 & 92.2 & 94.6 & 90.4 & 92.8 & 91.0 \\
Claude Opus 4.7 & 91.7 & 93.6 & 91.9 & 91.2 & 90.3 \\
GLM-5.1 & 91.5 & 93.6 & 91.0 & 90.2 & 91.3 \\
GPT-5.4 & 91.4 & 95.1 & 93.3 & 86.4 & 91.0 \\
Seed 2.0 Pro & 91.3 & 95.0 & 91.4 & 88.7 & 89.9 \\
DeepSeek V4 Pro & 90.8 & 94.0 & 89.5 & 90.3 & 89.6 \\
Qwen3.6 Max Preview & 90.6 & 91.7 & 91.9 & 89.3 & 89.5 \\
DeepSeek V4 Flash & 90.2 & 94.6 & 90.7 & 87.4 & 88.3 \\
Qwen3.5-27B & 89.8 & 92.9 & 92.2 & 86.0 & 88.3 \\
Claude Sonnet 4.6 & 89.4 & 95.2 & 85.0 & 87.7 & 89.7 \\
Qwen3.5-122B-A10B & 89.3 & 93.5 & 91.4 & 85.4 & 87.1 \\
Qwen3.5-35B-A3B & 89.3 & 92.9 & 90.7 & 86.2 & 87.4 \\
GPT-OSS-120B & 88.9 & 93.9 & 89.2 & 84.6 & 88.0 \\
Mimo v2.5 Pro & 88.8 & 94.8 & 88.5 & 85.4 & 86.5 \\
Qwen3.5-9B & 88.1 & 91.7 & 90.6 & 83.9 & 86.3 \\
GPT-OSS-20B & 85.6 & 91.9 & 85.9 & 82.1 & 82.3 \\
Llama-3.1-8B-Instruct & 51.6 & 49.7 & 51.3 & 53.4 & 52.2 \\
\midrule
Random & 50.0 & 50.0 & 50.0 & 50.0 & 50.0 \\
\bottomrule
\end{tabular*}

\vspace{0.35em}

\begin{tabular*}{0.82\textwidth}{@{\extracolsep{\fill}}lrrrrr@{}}
\toprule
\multicolumn{6}{c}{\ac} \\
\cmidrule(lr){1-6}
Model & Overall & Task & Planning & Tools & Rules \\
\midrule
GPT-5.4 & \textbf{89.4} & 86.4 & \textbf{92.6} & 89.7 & 88.9 \\
Gemini-3.1 Pro Preview & 86.5 & 84.7 & 83.7 & \textbf{90.3} & 87.3 \\
Claude Opus 4.7 & 85.0 & 86.5 & 82.7 & 81.8 & \textbf{89.0} \\
Kimi K2.6 & 84.3 & 82.7 & 81.4 & 85.8 & 87.3 \\
DeepSeek V4 Pro & 83.8 & 81.5 & 86.2 & 86.4 & 81.0 \\
Claude Sonnet 4.6 & 83.0 & 86.0 & 87.8 & 72.1 & 86.0 \\
DeepSeek V4 Flash & 81.3 & 86.7 & 81.4 & 74.4 & 82.8 \\
GLM-5.1 & 80.8 & \textbf{87.1} & 75.3 & 81.8 & 79.0 \\
Qwen3.6 Max Preview & 79.9 & 80.7 & 80.8 & 81.8 & 76.4 \\
Seed 2.0 Pro & 78.7 & 75.3 & 84.3 & 81.2 & 73.9 \\
Mimo v2.5 Pro & 78.5 & 81.4 & 78.2 & 78.3 & 76.1 \\
GPT-OSS-120B & 77.1 & 84.9 & 74.4 & 73.6 & 75.6 \\
Qwen3.5-122B-A10B & 75.0 & 78.0 & 74.4 & 73.8 & 73.9 \\
Qwen3.5-27B & 74.0 & 77.6 & 83.3 & 64.7 & 70.2 \\
GPT-OSS-20B & 72.3 & 80.0 & 80.1 & 59.2 & 69.8 \\
Qwen3.5-9B & 71.6 & 71.5 & 74.0 & 74.0 & 66.8 \\
Qwen3.5-35B-A3B & 70.8 & 73.8 & 75.6 & 61.4 & 72.4 \\
Llama-3.1-8B-Instruct & 61.7 & 56.8 & 62.8 & 72.0 & 55.2 \\
\midrule
Random & 50.0 & 50.0 & 50.0 & 50.0 & 50.0 \\
\bottomrule
\end{tabular*}
\caption{Main \benchmark leaderboard. Overall is \avgbacc; other columns are category \bacc. Scores are multiplied by 100 and shown with one decimal place. Models are ordered by Overall within each domain. Random denotes a uniformly random binary judge reference. Model sources: GPT-5.4~\citep{openai2026gpt54}, Gemini-3.1 Pro Preview~\citep{google2026gemini31pro}, Claude Opus 4.7~\citep{anthropic2026opus47}, Claude Sonnet 4.6~\citep{anthropic2026sonnet46}, Kimi K2.6~\citep{moonshot2026k2_6}, GLM-5.1~\citep{zai2026glm51}, DeepSeek V4 Pro~\citep{deepseek2026v4pro}, DeepSeek V4 Flash~\citep{deepseek2026v4flash}, Seed 2.0 Pro~\citep{bytedance2026doubaoseed20pro}, Qwen3.6 Max Preview~\citep{qwen2026max}, Qwen3.5 models~\citep{qwen3.5}, MiMo v2.5 Pro~\citep{mimo2026v25pro}, GPT-OSS-120B/20B~\citep{openai2025gptoss120bgptoss20bmodel}, and Llama-3.1-8B-Instruct~\citep{llama31modelcard}.}
\label{tab:main-leaderboard}
\end{table*}